%% file: anonymous-submission-latex-2024.tex
\documentclass[letterpaper]{article} 
\usepackage{aaai24}  
\usepackage{times}  
\usepackage{helvet}  
\usepackage{courier}  
\usepackage[hyphens]{url}  
\usepackage{graphicx} 
\usepackage{natbib}  
\usepackage{caption} 
\usepackage{amsmath}
\usepackage{enumitem}
\usepackage{algorithm}
\usepackage{algorithmic}
\usepackage{multirow}
\usepackage{amssymb}
\usepackage{newfloat}
\usepackage{listings}
\DeclareCaptionStyle{ruled}{labelfont=normalfont,labelsep=colon,strut=off} 
\floatstyle{ruled}
\newfloat{listing}{tb}{lst}{}
\floatname{listing}{Listing}
 
\title{SFANet: Spatial-Frequency Attention Network for \\ Weather Forecasting}

\author{
    Jiaze Wang\textsuperscript{\rm 1},
    Hao Chen\textsuperscript{\rm 1},
    Hongcan Xu\textsuperscript{\rm 1},
    Jinpeng Li\textsuperscript{\rm 1},
    Bowen Wang\textsuperscript{\rm 1},
    Kun Shao\textsuperscript{\rm 2},
    Furui Liu\textsuperscript{\rm 3},
    Huaxi Chen\textsuperscript{\rm 3},
    Guangyong Chen\textsuperscript{\rm 3},
    Pheng-Ann Heng\textsuperscript{\rm 1}
}
\affiliations{
    \textsuperscript{\rm 1}The Chinese University of Hong Kong\\
    \textsuperscript{\rm 1}Noah's Ark Lab, Huawei\\
    \textsuperscript{\rm 1}Zhejiang Lab\\

}

\usepackage{bibentry}

\begin{document}

\maketitle

\input{abstract}
\input{introduction}

\input{relatedwork}
\input{methodology}
\input{experiments}
\input{conclusion}

\clearpage

\bibliography{aaai24}

\end{document}

%% file: abstract.tex
\begin{abstract}

Weather forecasting plays a critical role in various sectors, driving decision-making and risk management. However, traditional methods often struggle to capture the complex dynamics of meteorological systems, particularly in the presence of high-resolution data. In this paper, we propose the Spatial-Frequency Attention Network (SFANet), a novel deep learning framework designed to address these challenges and enhance the accuracy of spatiotemporal weather prediction.
Drawing inspiration from the limitations of existing methodologies, we present an innovative approach that seamlessly integrates advanced token mixing and attention mechanisms. By leveraging both pooling and spatial mixing strategies, SFANet optimizes the processing of high-dimensional spatiotemporal sequences, preserving inter-component relational information and modeling extensive long-range relationships. To further enhance feature integration, we introduce a novel spatial-frequency attention module, enabling the model to capture intricate cross-modal correlations.
Our extensive experimental evaluation on two distinct datasets, the Storm EVent ImageRy (SEVIR) and the Institute for Climate and Application Research (ICAR) - El Niño Southern Oscillation (ENSO) dataset, demonstrates the remarkable performance of SFANet. Notably, SFANet achieves substantial advancements over state-of-the-art methods, showcasing its proficiency in forecasting precipitation patterns and predicting El Niño events.

\end{abstract}

%% file: introduction.tex

\section{Introduction}
Weather forecasting occupies a pivotal role within contemporary society, wielding extensive influence across diverse sectors such as agriculture, transportation, and disaster management~\cite{bi2023accurate,lam2022graphcast}. The precise anticipation of spatiotemporal weather dynamics assumes paramount importance, providing a foundational basis for informed decision-making, risk assessment, and judicious resource allocation. Notably, meteorological modeling has undergone significant evolution, yielding heightened forecast reliability. However, the dynamic and intricate nature of atmospheric phenomena, coupled with the burgeoning availability of high-resolution and multi-modal data, continues to pose persistent challenges to conventional forecasting methodologies~\cite{gneiting2005weather}.

In recent decades, the synergy of amplified computing power and improved data accessibility has catalyzed the development of intricate numerical models. These models encompass a comprehensive spectrum of atmospheric variables and physical processes, with General Circulation Models (GCMs)~\cite{weart2010development} and Regional Climate Models (RCMs)~\cite{rummukainen2010state} playing pivotal roles in weather prediction. While GCMs and RCMs empower researchers to simulate expansive climate patterns and regional weather phenomena, they grapple with challenges in accurately representing fine-scale attributes such as convective cells and localized wind patterns, which are crucial for short-term weather prognostications.
Moreover, the conventional practice of manual feature engineering and bespoke model design tailored to specific meteorological tasks faces constraints when confronted with the intricate and diverse tapestry of atmospheric data. 
Consequently, the field of meteorology is progressively embracing the realm of machine learning (Haupt et al., 2018; Bochenek and Ustrnul, 2022), leveraging its capabilities to unveil significant patterns and intricate relationships embedded within the abundance of accessible data.

Convolutional Neural Networks (CNNs)~\cite{guen2020disentangling,shi2015convolutional,shi2017deep,seo2023implicit} have showcased promise in decoding spatial patterns derived from weather radar and satellite imagery In meteorology. Concurrently, Recurrent Neural Networks (RNNs)~\cite{zaytar2016sequence,wang2017predrnn} have been harnessed to model temporal dependencies within sequential weather data, encompassing parameters like temperature and humidity time series. Nonetheless, these conventional CNNs and RNNs exhibit limitations in encapsulating both spatial and temporal attributes concurrently. The intricate interplay between diverse atmospheric factors, spanning wind direction, temperature gradients, and pressure fluctuations, necessitates elevated model sophistication to effectively encapsulate local interactions and far-reaching dependencies.

Guided by these challenges, we introduce an innovative method – the Spatial-Frequency Attention Network (SFANet) – meticulously crafted to elevate spatiotemporal weather forecasting by adeptly modeling intricate relationships interwoven within meteorological data. Our approach draws inspiration from recent strides in transformer-based architectures~\cite{vaswani2017attention} and token-level modeling~\cite{guibas2021efficient}. Transformers have highlighted their prowess in diverse sequence-to-sequence tasks, capitalizing on self-attention mechanisms to discern the salience of various components within sequences. SFANet advances this foundation, introducing two pioneering token mixing strategies: the pooling mixer~\cite{yu2022metaformer}  and the spatial mixer~\cite{guibas2021adaptive}.
The pooling mixer orchestrates information amalgamation across tokens, enabling selective emphasis on crucial features while mitigating noise. In parallel, the spatial mixer leverages the Fast Fourier Transform to encapsulate global attributes and far-reaching interdependencies – aspects often challenging for traditional models. This amalgamation empowers SFA to capture both local subtleties and overarching contextual insights, thereby augmenting the capacity to forecast intricate spatiotemporal meteorological patterns.
To further enhance feature fusion, we introduce a spatial-frequency attention (SFA) module within SFANet. This module adroitly captures correlations between spatial and frequency attributes, enriching the model's representation of meteorological data. By harmoniously melding spatial and frequency domains via the SFA module, forecasting precision attains an elevated threshold.

We conduct experiments on two weather forecasting datasets: SEVIR ~\cite{veillette2020sevir} dataset for precipitation nowcasting, and the ICAR-ENSO dataset~\cite{ham2019deep} for ENSO prediction. The results of our extensive experiments decisively underscore the preeminence of SFANet over contemporaneous methodologies across both datasets. SFANet yields substantial performance enhancements across the spectrum of evaluation metrics, thus affirming its proficiency in unraveling intricate spatiotemporal dynamics and prognosticating meteorological events.

Our main contributions can be summarized as follows:
\begin{itemize}[noitemsep,topsep=0pt]
\setlength{\itemsep}{0pt}
\item We introduce SFANet, a novel method for weather forecasting with pooling mixers and spatial mixers.
\item A spatial-frequency attention module is proposed to capture correlations between spatial and frequency attributes to generate more representative features.
\item Our results demonstrate that SFANet achieves significant improvements in both SEVIR benchmark for precipitation nowcasting and ICAR-ENSO benchmark for ENSO prediction. 
\end{itemize}

%% file: relatedwork.tex
\section{Related Works}
\noindent\textbf{Weather forecast modeling.}
Deep learning models for weather forecasting primarily rely on CNNs and RNNs due to their intrinsic capacity to enforce temporal and spatial inductive biases, facilitating the capture of complex spatiotemporal patterns. U-Net architectures ~\cite{ronneberger2015u,han2021convective,kwok2021enhanced}, employing either 2D or 3D CNNs~\cite{sun2022spatiotemporal,kim2022region}, have found application in diverse domains including precipitation nowcasting~\cite{veillette2020sevir,zhang2023skilful}, Seasonal Arctic Sea ice prediction~\cite{andersson2021seasonal}, and El Niño-Southern Oscillation (ENSO) forecasting\cite{ham2019deep}. Meanwhile, Long Short-Term Memory networks (LSTMs) ~\cite{hochreiter1997long} excel in modeling historical observation dynamics. This was underscored by Shi et al.~\cite{shi2015convolutional}, who extended the architecture to encompass convolutional structures within both input-to-state and state-to-state transitions, culminating in the Convolutional LSTM (ConvLSTM) designed for precipitation nowcasting. Subsequently, a rich body of research ~\cite{wang2018eidetic,wang2022predrnn,guen2020disentangling,ravuri2021skilful} has been dedicated to synergizing CNNs and RNNs for enhanced weather prediction, yielding substantial advancements in the field. Moreover, researchers ~\cite{keisler2022forecasting,lam2022graphcast} have explored the applicability of Graph Neural Networks (GNNs) for weather prediction. Recent developments ~\cite{chen2023fengwu,bi2023accurate} have ventured into the realm of transformer-based techniques for addressing Earth system forecasting challenges. Pathak et al. ~\cite{pathak2022fourcastnet} proposed FourCastNet, a global data-driven weather forecasting model that provides accurate short to medium-range global predictions.Bai et al. introduced Rainformer~\cite{bai2022rainformer}, a novel approach tailored to precipitation nowcasting. By ingeniously incorporating an efficient space-time attention block alongside a constellation of global vectors, Gao et al. introduced Earthformer~\cite{gao2022earthformer}, which attains state-of-the-art performance in Earth system forecasting. In this study, through an exhaustive exploration of cutting-edge weather forecasting paradigms, we present a novel transformer-based architecture that attains peak performance in weather forecasting tasks.

\noindent\textbf{Token mixing design.}
Transformers have demonstrated remarkable potential in the realm of computer vision tasks. A prevailing notion posits that their proficiency largely stems from the attention-based token mixer module. The original self-attention uses the similarity among the tokens to capture the long-range dependencies~\cite{vaswani2017attention,dosovitskiy2020image}. To balance the efficiency and performance, sparse attentions ~\cite{child2019generating,ho2019axial,beltagy2020longformer,parmar2018image} are proposed to promote predefined sparse patterns. Linformers ~\cite{wang2020linformer} and long-short transformers ~\cite{lian2021mlp} employ low-rank attention that utilize linear sketching to approximate self-attention. Kernel methods ~\cite{peng2021random,katharopoulos2020transformers,choromanski2020rethinking} and clustering-based approaches ~\cite{roy2021efficient,zaytar2016sequence,tay2020sparse} have also been harnessed to approximate attention mechanisms.
To relax the graph similarity constraints of the self-attention, Tolstikhin et al. ~\cite{tolstikhin2021mlp} originally proposed MLP-mixer using MLP projections that achieves similar accuracy as self-attention. It is further accelerated by ResMLP ~\cite{touvron2022resmlp} and gMLP ~\cite{liu2021pay}. The Fourier transform has also been attempted to spatially mix tokens. For instance, FNet ~\cite{lee2021fnet} resembles the MLP-mixer with simply pre-fixed DFT token mixer. Guibas et al. designed Adaptive Fourier Neural Operators (AFNO)~\cite{guibas2021adaptive} by imposing block-diagonal structure, adaptive weight-sharing, and sparsity. Employing a simple non-parametric operator, pooling, as an extremely weak token mixer, Yu et al. ~\cite{yu2022metaformer} build a simple model named PoolFormer and find it can still achieve highly competitive performance.

%% file: methodology.tex
\section{Method}
\begin{figure}[t]
    \centering
        \includegraphics[width=\linewidth]{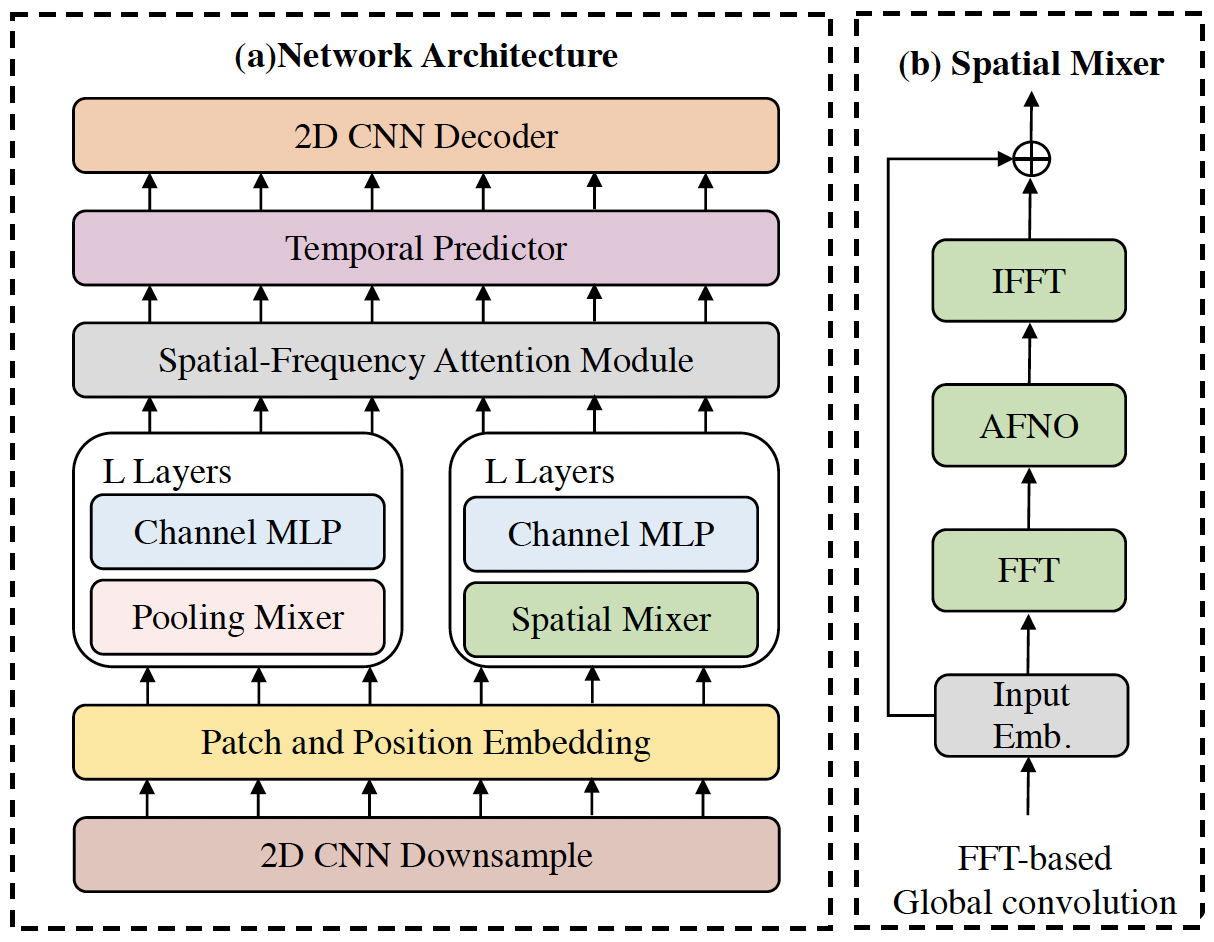}
    \caption{\textbf{Illustration of the SFANet architecture.} (a) Overview of the network.  We adopt the encoder-predictor-decoder architecture for weather forecasting. The encoder consists of a 2D CNN for downsampling, $L$ pooling mixer blocks to generate features at the spatial domain, $L$ spatial mixer blocks to generate features at the frequency domain, and a spatial-frequency attention module to capture the correlations between spatial and frequency features. (b) The process of our spatial mixer. (\romannumeral1) Spatial token mixing via fast Fourier transform. (\romannumeral2) Adaptive Fourier Neural Operator to mix in the Fourier domain. (\romannumeral3) Inverse FFT for token demixing.}
    \label{fig:overview}
    \vspace{-12pt}
\end{figure}
In this section, we first provide a brief review of the spatiotemporal sequences forecasting problem. Subsequently, we introduce an innovative token mixer module that harnesses the capabilities of both the pooling mixer and spatial mixer, effectively capturing extensive long-range relationships. In addition, we present a spatial-frequency attention module designed to capture the inherent correlations between spatial and frequency features. Complementing these modules, we incorporate a predictor responsible for encoding temporal information, as well as a decoder tasked with reconstructing future sequences.

\subsection{Preliminary}
\noindent\textbf{Problem statement.}
Similar to previous works~\cite{shi2015convolutional,gao2022earthformer}, we formulate weather forecasting as a spatiotemporal sequences forecasting problem, which aims to infer future spatiotemporal sequences using the previous ones. Given a spatiotemporal sequence $[X_i]_{i=1}^T$ at time $t$ with the past $T$ frames, our goal is to predict the future sequence $[Y_{T+i}]_{i=1}^K$ at time $t$ that contains the next $K$ frames, where $X_{i}\in \mathbb{R}^{H \times W \times C}$. Here $H$ denotes image height, $W$ denotes image width, and $C$ denotes image channels. The target of the task is to train a mapping function $F_\Theta$ with the learnable parameters $\Theta$ to minimize the following loss function:

\begin{align}
\Theta=\mathop{\arg\min}_{\Theta} L([X_i]_{i=1}^T,[Y_{T+i}]_{i=1}^K)
\end{align}

where L can be various loss functions, such as mean squared error(MSE), mean absolute error (MAE) or smooth loss. In this work, we employ MSE loss in our setting as many previous sequences forecasting works do~\cite{gao2022simvp}.

\subsection{Token Mixer}
\textbf{Motivation.} 
Divergent from convolutional neural networks, architectures employing the self-attention mechanism, such as Vision Transformers (ViT)~\cite{dosovitskiy2020image}, possess a lesser degree of inductive bias, necessitating a heightened dependence on information acquisition from data. Notably, for high-resolution tasks like weather prediction, employing larger patch sizes can lead to a loss of inter-component relational information, while opting for smaller patch sizes introduces concerns of the predicted sequences. Furthermore, the significance of long-range relationships cannot be well modeled. The inter-token relationships are profoundly influenced by the chosen mixing methodology. Thus, we adopt two kinds of advanced token-mixing approaches to address these challenges. More details are given in the supplementary material.

\noindent\textbf{Pooling Mixer.}
As is widely acknowledged, self-attention and spatial MLP operations exhibit a computational complexity that grows quadratically with the number of tokens being processed. This issue is further exacerbated in the case of spatial MLPs, which introduce a substantial increase in parameters when dealing with longer sequences. Consequently, self-attention and spatial MLPs tend to be limited to processing sequences of limited tokens.
In sharp contrast, pooling operations offer a distinctive advantage by maintaining a computational complexity that scales linearly with the length of the sequence. Remarkably, pooling achieves this efficiency without the incorporation of any additional learnable parameters.
Therefore, we simply use the pooling operator as the mixer which has no learnable parameters and just makes each token averagely aggregate its nearby token features. We denote the input feature embedding of the pooling layer as $F\in \mathbb{R}^{H \times W \times C}$, the pooling size is $K$, the pooling operator can be calculated as:
\begin{align}
F^{'}_{i,j,:} = \frac{1}{K^2} \sum_{m,n = 1}^{k} F^{'}_{i+m-\frac{K+1}{2} ,j+n-\frac{K+1}{2},:}-F_{i,j,:}
\end{align}

\noindent\textbf{Spatial Mixer.} 
Expanding upon the foundational principles introduced by AFNO~\cite{guibas2021efficient}, we advance our methodology by extending token mixing to encompass the Fourier domain. This innovation facilitates the harmonious fusion of diverse modalities inherent in the Fourier spectrum. Leveraging the Fourier domain inherently capitalizes on its global attributes, thereby effectively mitigating concerns pertaining to long-range relationships. To begin, in addressing the requirement for efficient global convolution, we incorporate spatial token mixing via the fast Fourier transform (FFT).

Given the intrinsic high-resolution nature of weather sequences, characterized by intricate nuances, we acknowledge that Fourier Neural Operator methodologies introduce a computational complexity that escalates quadratically with the channel size. To adeptly manage and alleviate this computational overhead, we ingeniously introduce a block-diagonal structure, seamlessly integrated into the channel mixing weights.
In addition, to augment the model's aptitude for generalization, we introduce a soft-thresholding strategy to inculcate sparsity within the frequency components. This strategic maneuver augments the model's flexibility in handling diverse scenarios.
Furthermore, with an emphasis on optimizing parameter efficiency, we employ a weight-sharing mechanism across tokens within our Multi-Layer Perceptron (MLP) layer. This judiciously executed approach streamlines the utilization of parameters, thus enhancing the model's capacity to astutely learn and effectively represent intricate relationships.
By methodically amalgamating these refined components, we fortify our framework's aptitude to comprehend and encapsulate intricate spatiotemporal dependencies within meteorological data, thereby paving the way for enhanced forecasting performance.

\begin{table*}[]
\centering
\resizebox{\linewidth}{!}{%
\begin{tabular}{c|c|c|cccccccc}
\hline
\multirow{2}{*}{Model} & \multirow{2}{*}{Param.(M)} & \multirow{2}{*}{GFLOPS} & \multicolumn{8}{c}{Metrics} \\
 &  &  & CSI-M & CSI-219 & CSI-181 & CSI-160 & CSI-133 & CSI-74 & CSI-16 & MSE \\ \hline
Persistence & - & - & 0.2613 & 0.0526 & 0.0969 & 0.1278 & 0.2155 & 0.4705 & 0.6047 & 11.5338 \\
ConvLSTM~\shortcite{shi2015convolutional} & 14.0 & 528 & 0.4185 & 0.1288 & 0.2482 & 0.2928 & 0.4052 & 0.6793 & 0.7569 & 3.7532 \\
E3D-LSTM~\shortcite{wang2018eidetic} & 35.6 & 523 & 0.4038 & 0.1239 & 0.2270 & 0.2675 & 0.3825 & 0.6645 & 0.7573 & 4.1702 \\
Unet~\shortcite{veillette2020sevir}  & 16.6 & 33 & 0.3592 & 0.0577 & 0.1580 & 0.2157 & 0.3274 & 0.6531 & 0.7441 & 4.1119 \\
PhyDNet~\shortcite{guen2020disentangling} & 13.7 & 701 & 0.3940 & 0.1288 & 0.2309 & 0.2708 & 0.3720 & 0.6556 & 0.7059 & 4.8165 \\
PredRNN~\shortcite{wang2022predrnn} & 46.6 & 328 & 0.4080 & 0.1312 & 0.2324 & 0.2767 & 0.3858 & 0.6713 & 0.7507 & 3.9014 \\
Rainformer~\shortcite{bai2022rainformer} & 184.0 & 170 & 0.3661 & 0.0831 & 0.1670 & 0.2167 & 0.3438 & 0.6585 & 0.7277 & 4.0272 \\
Earthformer~\shortcite{gao2022earthformer}  & 15.1 & 257 & 0.4419 & 0.1791 & 0.2848 & 0.3232 & 0.4271 & 0.6860 & 0.7513 & 3.6957 \\ \hline
Ours & 16.4 & 185 &  \textbf{0.4692}  & \textbf{0.1820} & \textbf{0.2905} & \textbf{0.3526} & \textbf{0.4658} & \textbf{0.7356} & \textbf{0.7887} & \textbf{2.7308} \\ \hline
\end{tabular}
}
\caption{\textbf{Comparison with state-of-the-art methods on the SEVIR}.}
\label{table:sevir}
\vspace{-6pt}
\end{table*}

\subsection{Model architecture}

\noindent\textbf{Encoder-Predictor-Decoder Design.}
As illustrated in Figure \ref{fig:overview}, we adopt the encoder-predictor-decoder model architecture for weather forecasting. The encoder extract features from the observed sequences and forward the features to the predictor, then the predictor predicts future features at the feature level. Finally, the decoder decodes the future feature to reconstruct the future sequences.

\noindent\textbf{Spatial-Frequency Attention Module.}
To model the inherent relation of the feature in the spatial space and frequency space, the spatial-frequency attention module is proposed.
Capturing the correlations between spatial and frequency attributes assumes significance in enhancing the model's representation of meteorological data.
Given these properties, we propose to capture their relations and deeply integrate them using a relation function $\mathcal{R}$. In this way, we reproduce the relation-injected spatial feature and frequency feature of the instance as follows:
\begin{equation}
    \hat{F_s} = F_s + \mathcal{R}(F_s, F_f), \quad
    \hat{F_f} = F_f + \mathcal{R}(F_f, F_s).
\end{equation}
We add the original feature to the relation feature, in order to drive the network to unearth as much complementary information as possible by relational learning. Then we concatenate $\hat{F_s}$ and $\hat{F_f}$ as our relation-enhanced encoder embedding $F_i$ to the predictor. 
We leverage three prominent and widely recognized architectures: Multi-Layer Perceptron (MLP)\cite{santoro2017simple}, Cross Attention\cite{huang2019ccnet}, and Transformer~\cite{devlin2018bert}. Our empirical investigations substantiate that the lightweight Transformer architecture adeptly captures these relationships.

\noindent\textbf{Temporal Predictor.}
We leverage the predictor's capability to encode the input sequence across multiple levels of representation, employing a coarse-to-fine approach to generate predictions.
State-of-the-art predictors in the field can be broadly categorized into two main types: RNN-based and CNN-based predictors. The former utilizes recurrent neural networks to project video frames into a latent space, subsequently utilizing the RNN architecture to forecast future latent states. This approach has garnered significant attention due to its capacity to extract concise yet informative features, resulting in accurate and efficient predictions. On the other hand, the latter framework revolves around the simplicity and effectiveness of CNN-based methods. For instance, SimVP~\cite{gao2022simvp} introduces an efficient CNN-based predictor that capitalizes on Inception modules to capture temporal evolution. The Inception module comprises a bottleneck Conv2d layer with a 1×1 kernel, followed by parallel GroupConv2d operators.
We conducted a comprehensive experimental analysis encompassing both RNN-based and CNN-based predictors. Remarkably, our findings highlight the superior performance of the CNN-based predictor proposed in SimVP in the context of weather forecasting tasks.

\noindent\textbf{Decoder}
The purpose of the decoder is to reconstruct future sequences, which is $C$ channels on $(H, W)$. We adopt the non-auto-regressive decoder to avoid the accumulating error problem and increase the model efficiency. The decoder is composed $N$ blocks, each block consists of a ConvTranspose2d layer, a GroupNorm layer and a LeakyReLU layer.

%% file: experiments.tex
\section{Experiments}

\subsection{Experimental Setup}

\noindent \textbf{SEVIR Precipitation Nowcasting.}
The Storm EVent ImageRy (SEVIR) dataset comprises over 10,000 annotated and spatio-temporally aligned weather events, with each event containing 384 km × 384 km image sequences spanning 4 hours. The dataset includes five data types, such as three channels (C02, C09, C13) from GOES-16 advanced baseline imager, NEXRAD Vertically Integrated Liquid (VIL) mosaics, and GOES-16 Geostationary Lightning Mapper (GLM) flashes. SEVIR events are matched with the NOAA Storm Events database, enabling additional descriptive information linkage. It serves as a valuable benchmark for meteorological applications, and we utilize it for precipitation nowcasting experiments, predicting future VIL values within 60 minutes given 65 minutes of contextVIL observations.

\noindent\textbf{ICAR-ENSO Sea Surface Temperature Anomalies Forecasting.}
The ICAR-ENSO dataset, provided by the Institute for Climate and Application Research (ICAR), encompasses historical climate observations and simulations. It enables the forecasting of Sea Surface Temperature (SST) anomalies up to 14 steps, allowing for predictions beyond one year with a context of 12 steps of SST anomaly observations. This dataset combines observational and simulation data to offer forecasts of El Niño/Southern Oscillation (ENSO), a critical indicator of seasonal climate worldwide due to its influence on sea surface temperatures in the Equatorial Pacific

\noindent\textbf{Metrics.}
For rain forecasting models, we employ the Critical Success Index (CSI) as a key evaluation metric. Widely utilized in precipitation nowcasting, CSI is formally expressed as $CSI=Hits/Hits+Misses+F.Alarms$. It quantifies $Hits$, $Misses$, and $False Alarms$ after rescaling predictions and ground truth to a 0-255 range and applying binarization at specific thresholds. Our report encompasses CSI scores at different thresholds [16, 74, 133, 160, 181, 219], along with their mean CSI-M.
Furthermore, our investigation validates ENSO forecasting via Nino SST indices~\cite{gao2022earthformer}. Particularly, the Nino3.4 index reflects averaged anomalies in Sea Surface Temperatures (SST) within a defined Pacific region (170°W-120°W, 5°S-5°N). This index plays a pivotal role in identifying El Nino events, leveraging SST anomalies along the equator.

\noindent\textbf{Implementation Details.}
We use the AdamW optimizer with $\beta_1$ = 0.9 and $\beta_2$ = 0.999, and we incorporate a 20\% linear warm-up and a cosine learning rate scheduler, which gradually reduces the learning rate from its maximum to zero after the warm-up phase. The training spans 100 epochs across all datasets, and we implement early stopping based on the validation score, with a tolerance of 20. We use 4 A100 GPUs for training and 1 A100 for evaluation.

\subsection{Results}

\noindent\textbf{Results on SEVIR.}
The outcomes pertaining to SEVIR are presented in Table \ref{table:sevir}. SFANet consistently showcases a remarkable performance superiority when juxtaposed against the established baseline methodologies, encompassing a diverse spectrum of metrics. SFANet attains notable achievements, with a recorded Mean Squared Error (MSE) of 2.7308, CSI-16 of 0.7787, and CSI-219 of 0.1820. This substantial advancement stands as a noteworthy departure from previous methods, signifying a remarkable stride in performance improvement. Evidently, SFANet demonstrates its efficacy across a range of conditions, affirming its superior adaptability.
Remarkably, the discernible performance enhancement is achieved while maintaining comparable model parameters and, in some instances, even fewer Giga-Floating Point Operations Per Second (GFLOPS). This dual accomplishment accentuates the profound contributions and innovative advancements that SFANet introduces in ameliorating the inherent challenges intrinsic to the task.

\noindent\textbf{Results on ICAR-ENSO.}
As evidenced in Table \ref{table:icar}, our approach consistently outperforms all alternative methods across the entire spectrum of evaluated metrics. Specifically, we observe enhancements of 4.7\% on C-Nino3.4-M, 8.2\% on C-Nino3.4-WM, and a reduction of 3.5\% in Mean Squared Error (MSE). These substantial improvements corroborate the robust effectiveness of our methodology within the realm of ESNO prediction.

\begin{table}[]
\centering
\resizebox{\linewidth}{!}{%
\begin{tabular}{c|ccc}
\hline
\multirow{2}{*}{Model}  & \multicolumn{3}{c}{Metrics} \\
 & C-Nino3.4-M & \multicolumn{1}{c}{C-Nino3.4-WM} & \multicolumn{1}{c}{MSE} \\ \hline
Persistence  & 0.3221 & 0.447 & 4.581 \\
ConvLSTM~\shortcite{shi2015convolutional} & 0.6955 & 2.107 & 2.657 \\
E3D-LSTM~\shortcite{wang2018eidetic}  & 0.7040 & 2.125 & 3.095 \\
Unet~\shortcite{veillette2020sevir}  & 0.6926 & 2.102 & 2.868 \\
PhyDNet~\shortcite{guen2020disentangling}  & 0.6646 & 1.965 & 2.708 \\
PredRNN~\shortcite{wang2022predrnn}  & 0.6492 & 1.910 & 3.044 \\
Rainformer~\shortcite{bai2022rainformer}  & 0.7106 & 2.153 & 3.043 \\
Earthformer~\shortcite{gao2022earthformer} & 0.7329 & 2.259 & 2.546 \\ \hline
Ours & \textbf{0.7689} & \textbf{2.445} & \textbf{2.457} \\ \hline
\end{tabular}
}
\caption{\textbf{Comparison with state-of-the-art methods on the ICAR-ENSO}.
}
\label{table:icar}
\vspace{-10pt}
\end{table}

\begin{figure*}[t]
    \centering
        \includegraphics[width=\linewidth]{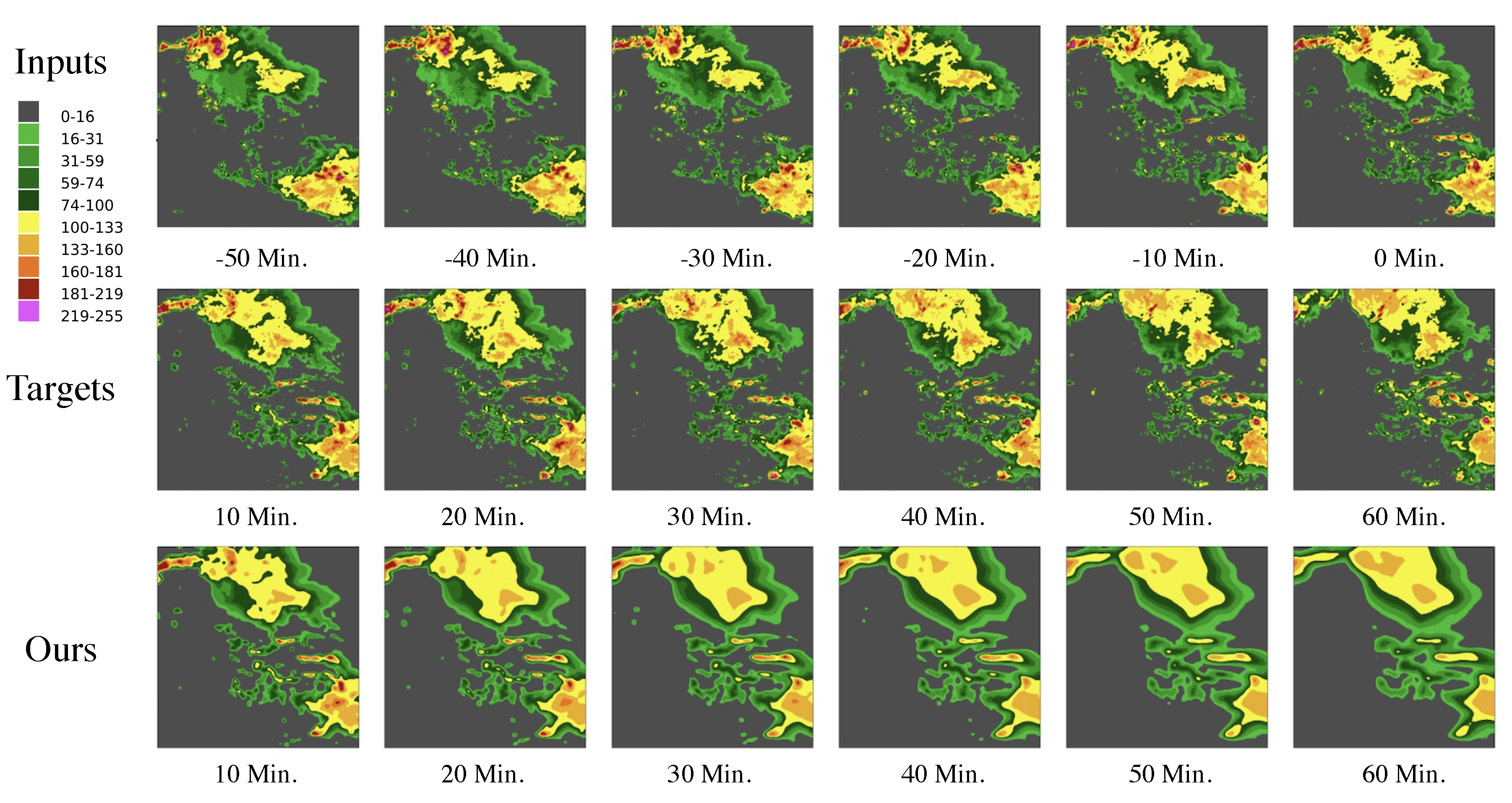}
    \caption{\textbf{Prediction results of SFANet on the SEVIR dataset.} Illustrated via vertically integrated liquid water contents, quantified on a scale ranging from 0 to 255 shown in the color bar.}
    \label{fig:vis1}
    \vspace{-4pt}
\end{figure*}
\subsection{Ablation Study}
We have undertaken an ablation study utilizing the SEVIR dataset, aimed at dissecting the impact of each component constituting SFANet. The findings from this study are meticulously detailed in Table \ref{table:ablation}, wherein we systematically evaluate the performance of our methodology by progressively integrating individual components onto the baseline.

\begin{table*}[]
\centering
\resizebox{\linewidth}{!}{%
\begin{tabular}{c|c|c|cccccccc}
\hline
\multirow{2}{*}{Token   Mixer} & \multirow{2}{*}{Predictor} & \multirow{2}{*}{SFA Module} & \multicolumn{8}{c}{Metrics} \\
 &  &  & CSI-M & CSI-219 & CSI-181 & CSI-160 & CSI-133 & CSI-74 & CSI-16 & MSE \\ \hline
Pooling & SimVP & - & 0.4110 & 0.1111 & 0.2274 & 0.2769 & 0.3942 & 0.6918 & 0.7645 & 3.5872 \\
Spatial & SimVP & - & 0.4145 & 0.1302 & 0.2448 & 0.2784 & 0.4208 & 0.6792 & 0.7335 & 3.8750 \\
Pooling+Spatial & SimVP & - & 0.4433 & 0.1394 & 0.2707 & 0.3196 & 0.4344 & 0.7201 & 0.7756 & 3.2694 \\
Pooling+Spatial & PhyDNet & - & 0.4263 & 0.1338 & 0.2523 & 0.2884 & 0.4195 & 0.6998 & 0.7641 & 3.4221 \\
Pooling+Spatial & SimVP & MLP & 0.4537 & 0.1590 & 0.2862 & 0.3326 & 0.4464 & 0.7217 & 0.7760 & 2.9864 \\
Pooling+Spatial & SimVP & CrossAttention & 0.4540 & 0.1513 & 0.2874 & 0.3351 & 0.4467 & 0.7247 & 0.7788 & 2.9052 \\
Pooling+Spatial & SimVP & Transformer & \textbf{0.4692} & \textbf{0.1820} & \textbf{0.2905} & \textbf{0.3526} & \textbf{0.4658} & \textbf{0.7356} & \textbf{0.7887} & \textbf{2.7308} \\ \hline
\end{tabular}
}
\caption{\textbf{Ablation Study of different token mixer, predictor and SFA Module on SEVIR.} }
\label{table:ablation}
\end{table*}

\noindent\textbf{Token mixer}
In contrast to the attention module typically employed in Transformer architectures, our methodology introduces a pivotal modification through the integration of two advanced token mixers. Ablation studies commence with an assessment of the straightforward pooling mixer. The results substantiate the efficacy of this approach, with our method achieving a competitive CSI-16 score of 0.7645, thus reinforcing the assertion that the pooling mixer serves as a potent yet uncomplicated token mixer for the task at hand.
Subsequently, the pooling mixer is supplanted by the spatial mixer, leveraging the expeditious Fast Fourier Transform. While the spatial mixer exhibits diminished performance in terms of CSI-16 and MSE, it compensates by yielding superior outcomes in more challenging assessment metrics such as CSI-219 and CSI-181. This affirms the spatial mixer's adeptness in capturing high-resolution content characterized by intricate details.
Notably, a more comprehensive enhancement materializes when both token mixers are synergistically integrated. This substantiates the symbiotic relationship between the distinct feature embeddings encapsulated within the two mixers. The resulting superior performance underscores the compelling complementarity between these diverse characteristics, culminating in an improved predictive prowess.

\noindent\textbf{Predictor Selection}
As investigated by previous works, CNN and RNN based methods achieve state-of-the-art performance on sequence prediction. Notably, these approaches offer distinct advantages: RNNs exhibit swifter convergence when endowed with ample model capacity, while CNNs boast robust training dynamics and exhibit minimal fluctuations even at elevated learning rates. To comprehensively assess their efficacy, we undertake an ablation study that encompasses both CNN and RNN-based predictors. In the realm of CNN-based prediction, we adopt the Inception-based predictor proposed in SimVP. Conversely, for the RNN-based counterpart, we employ PhyCell~\cite{guen2020disentangling}, a novel temporal module that effectively accounts for physical dynamics during modeling. Our empirical findings affirm the superiority of utilizing the predictor introduced in SimVP, as it consistently attains enhanced performance across the entire gamut of evaluation metrics.

\noindent\textbf{SFA Module.}
Our investigation substantiates that each of the MLP, CrossAttention, and Transformer architectures yields an enhancement compared to the model devoid of the Spatial-Frequency Attention (SFA) module. This discernible progress underscores the crucial role played by the SFA module in effectively amalgamating spatial and frequency features. Furthermore, noteworthy observations arise from our investigation, particularly highlighting the remarkable efficacy of the Transformer-based architecture. This particular structure consistently yields the most favorable outcomes across all evaluation metrics. This compelling pattern underscores the preeminence of our adeptly designed lightweight Transformer modules in effectively capturing intricate long-range and cross-modal feature relationships.

\begin{figure*}[t]
    \centering
        \includegraphics[width=\linewidth]{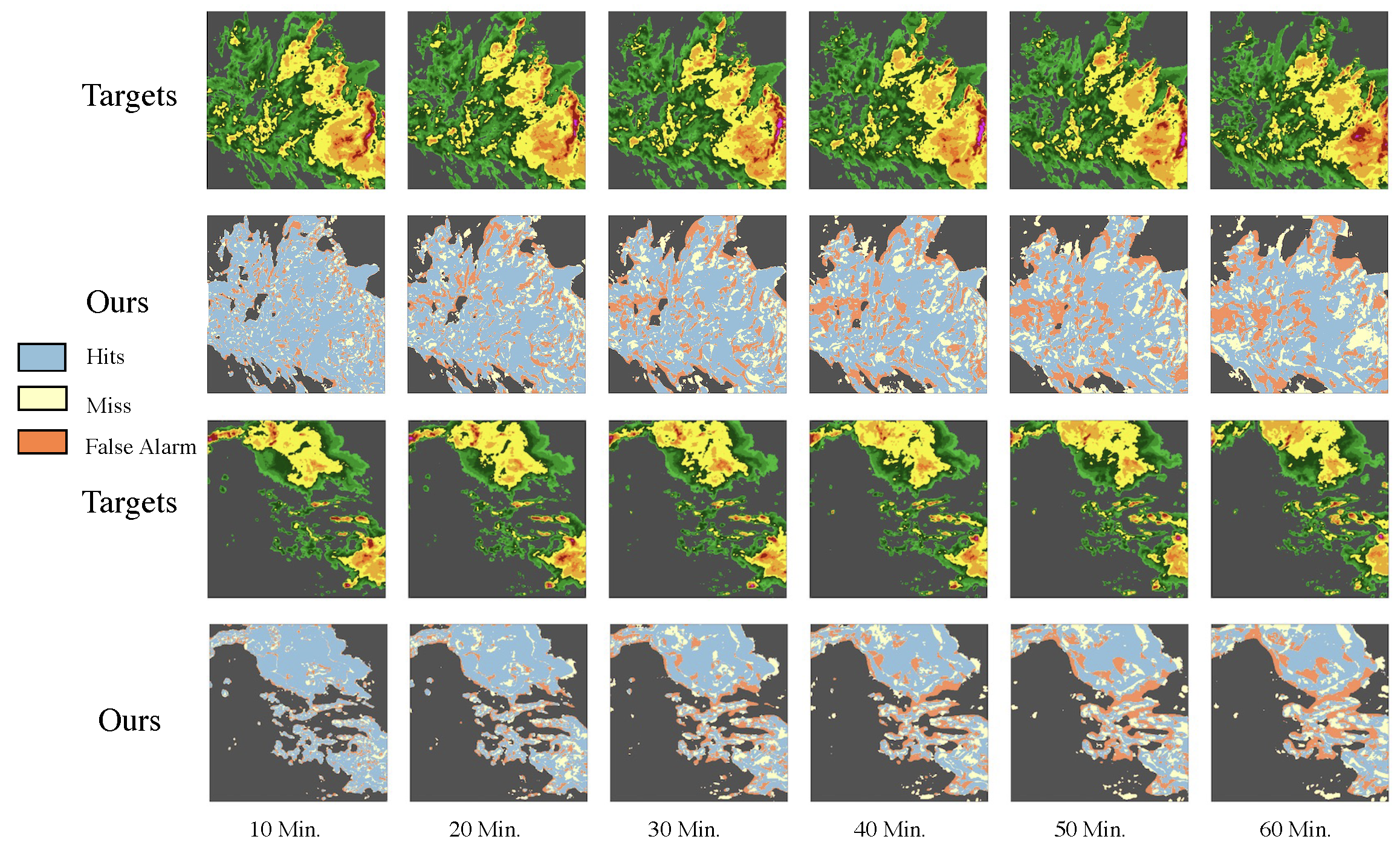}
    \caption{\textbf{Error Analysis of SFANet on the SEVIR dataset.} For each pixel in prediction results (Predict) and ground truth (GT), we will assign it a label according to the thresholds ([16, 74, 133, 160, 181, 219]). Hits: $Predict=GT$. Miss: $Predict<GT$. False Alarm: $Predict>GT$.}
    \label{fig:vis2}
    \vspace{-10pt}
\end{figure*}
\subsection{Qualitative Analysis.}

\noindent\textbf{Prediction results visualization.} In our visual analysis, depicted in Figure \ref{fig:vis1}, we initially observe that over time, the predicted Vertically Integrated Liquid (VIL) data becomes smoother. This suggests that forecasting VIL for distant timeframes presents greater challenges, warranting further in-depth investigation. Furthermore, we note that our predictive outcomes tend to underestimate regions with heavier precipitation intensity. We attribute this tendency to the scarcity of data instances with higher values within the dataset. Consequently, the model exhibits a bias towards predicting lower levels of precipitation intensity than actual observations. Given that heavy precipitation events often have a more pronounced impact on daily life, our future research should emphasize accurate prediction of such weather patterns to enhance the model's practical utility.

\noindent\textbf{Error analysis visualization.} In Figure \ref{fig:vis2}, we analyze the Hits, Misses, and False Alarms for two sequences. It becomes apparent that the model's errors accumulate as time progresses, indicating that long-term weather forecasting presents a greater challenge. This observation aligns with our earlier analysis in Figure \ref{fig:vis2}. Additionally, we observe a tendency for prediction errors to propagate from previously mispredicted areas to their surrounding regions over time. Addressing the issue of cumulative model errors is a significant avenue for future research, warranting focused investigation.

%% file: conclusion.tex

\section{Conclusion}
In this study, we introduce SFANet, a novel framework for accurate spatiotemporal weather forecasting. SFANet utilizes advanced token mixing mechanisms, a spatial-frequency attention module, and an encoder-predictor-decoder architecture to capture long-range dependencies and cross-modal correlations in weather data. Our approach addresses challenges in high-resolution data and long-range relationships by integrating pooling and spatial token mixers. The spatial-frequency attention module effectively combines spatial and frequency features, enhancing cross-modal correlations. The model's encoder-predictor-decoder architecture encodes input sequences, predicts future features, and reconstructs sequences.
SFANet's efficacy is validated through comprehensive experiments on SEVIR and ICAR-ENSO datasets, demonstrating superior performance compared to state-of-the-art methods. The model achieves remarkable results in all evaluation metrics, showcasing its adaptability across different thresholds. Our qualitative analysis indicates SFANet's ability to capture evolving weather patterns.

Though impressive results have achieved, SFANet still has some limitations. Firstly, there exists potential for refinement in accurately forecasting heavy precipitation, leaving an avenue for further exploration in future endeavors. Additionally, challenges related to blurriness and the accumulation of errors persist, as is often observed in deep learning based models. These issues remain subjects of ongoing investigation, and we will investigate them in the future.